\documentclass[twoside,11pt]{article}
\usepackage[english]{babel}
\usepackage{enumerate}
\usepackage{amssymb,amsfonts,amsmath,latexsym}
\usepackage{amsthm}
\usepackage{graphicx}

\usepackage{caption}

\usepackage[titletoc,title]{appendix}

\usepackage{geometry}
\usepackage{color}
\usepackage{array}
\usepackage{float}
\usepackage{multirow}
\usepackage{graphicx}
\usepackage{setspace}
\usepackage{wasysym}
\usepackage{amssymb,amsfonts,amsmath,latexsym}
\usepackage{amsthm}
\usepackage{etex,etoolbox}
\usepackage{amsthm,amssymb}
\usepackage{thmtools}
\usepackage{environ}
\usepackage{jair, theapa, rawfonts}

\begin{document}
\title{De-Biasing Models of Biased Decisions: \\ A Comparison of Methods Using Mortgage Application Data}

\author{\name Nicholas H. Tenev\thanks{The views expressed in this paper are my own and do not represent the views of the OCC, the Department of the Treasury, or the United States government.} \email nicholas.tenev@occ.treas.gov\\
	\addr Office of the Comptroller of the Currency\\ 400 7th St SW, Washington, DC 20219}
\date{July 12, 2023}
\maketitle
\begin{abstract}
Prediction models can improve efficiency by automating decisions such as the approval of loan applications. However, they may inherit bias against protected groups from the data they are trained on. This paper adds counterfactual (simulated) ethnic bias to real data on mortgage application decisions, and shows that this bias is replicated by a machine learning model (XGBoost) even when ethnicity is not used as a predictive variable. Next, several other de-biasing methods are compared: averaging over prohibited variables, taking the most favorable prediction over prohibited variables (a novel method), and jointly minimizing errors as well as the association between predictions and prohibited variables. De-biasing can recover some of the original decisions, but the results are sensitive to whether the bias is effected through a proxy.
\end{abstract}



\section{Introduction}\label{sec:intro}

Prediction models can be used to automate decisions previously made by people. But if a model is trained to replicate historical decisions that were biased against certain prohibited basis groups (``PBGs'') such as racial or ethnic groups, the model predictions may perpetuate that bias. Even if group membership is not used as a predictive variable in the model, the model can learn to weight the predictive variables it does use in a way that replicates the bias. This paper demonstrates using real data that this can occur in the context of mortgage underwriting, and then studies the performance of several solutions capable of removing some of the bias, including one (maximum over prohibited groups) that is new to the literature.

Many of the relevant inputs to making an underwriting decision are standardized and quantifiable, making approving loan applications a candidate for automation. To do so, a lender might create a model which mimics historical underwriting decisions made by loan officers. Another application of this paper might be a lender or regulatory agency modeling underwriting decisions as a baseline against which to identify outliers, for quality control or as part of fair lending monitoring. This paper studies the case of modeling approval decisions rather than modeling loan performance, avoiding the sample selection challenge that occurs in trying to make inferences about performance when it is only observed for approved applicants.\footnote{\citeA{grau2021pbot} and \citeA{hull2021outcome} study how outcomes can be used to overcome this selection issue. The methods studied in this paper may also be useful in de-biasing models trained on performance data (e.g. mortgage delinquencies rather than mortgage approvals)---future work will explore this possibility.} 

First, this paper demonstrates that a model trained on historical decision data can replicate bias against a group even if group membership is not used as a predictive variable. Simulated bias against Hispanic and Latino applicants is added to data on mortgage application decisions, and a machine learning model is trained on the biased decisions using only standard underwriting variables (but not borrower ethnicity). While the bias added is artificial, the data are otherwise real---this demonstrates that in the context of mortgage underwriting in the US, the correlations between prohibited basis group and other predictive variables are strong enough to allow a machine learning model to replicate bias even when it does not explicitly use the group as a predictor.\footnote{Whether or not a given model will actually replicate bias in the decisions it is being trained to predict is an empirical question, depending on both the data and the model in question. Some models may not be flexible enough to discriminate between certain groups, and some groups may not differ sufficiently in the predictive variables for even the most flexible model to tell them apart.}

Next, the paper tests several methods of removing bias: averaging over the protected group (adapting the technique of \citeR{pope2011implementing} to the machine learning context), regularization to reduce the association between protected group and outcome (FairXGBoost---\citeR{ravichandran2020fairxgboost}), and a novel method: taking the most favorable prediction over protected groups. In this last method, a model is trained using all available information (including prohibited characteristics), and then de-biased by changing the prediction from deny to approve for any case in which a prohibited variable was the deciding factor (see Section \ref{sec:lop} for more detail). None of these de-biasing methods results in a model which uses a PBG as a predictive variable. That is, two applicants with the same credit score, debt-to-income ratio, etc. will receive the same prediction from a given model regardless of their race or ethnicity. Prohibited basis group is only used for de-biasing, to determine how much the model should depend on other predictive variables. The legality of using prohibited basis group information, even for de-biasing, is beyond the scope of this paper. 

The literature on bias mitigation in machine learning and statistics is vast; \citeA{hort2022bia} survey the former. Some proposals involve ``pre-processing'' of the training data, such as modifying the training data, as in \citeA{classwithout} or \citeA{verma2021removing}, re-weighting the training data\footnote{In this case, the training data consist of the historical loan applications and loan officer decisions used to train the model.}, as in \citeA{buildingclass2009}, or creating a ``fair representation'' of the data for training that loses information about the protected group while retaining other information, as in \citeA{learningfair2013}. By contrast, ``in-processing'' approaches alter the objective of training. One way to do this is a ``regularization'' approach, jointly optimizing the accuracy of the predictions\footnote{The FairXGBoost method studied in this paper measures accuracy as cross-entropy loss between the model predictions and the actual data.} and a measure of disparity between groups, such as the FairXGBoost algorithm tested in this paper (\citeR{ravichandran2020fairxgboost}) or \citeA{fairnessaware2011}. Another is to add a constraint that cannot be breached during optimization, as in \citeA{constraint2019}. ``Post-processing'' methods mitigate bias by altering the model or predictions once it has been trained, e.g. \citeA{threenaive2010}. Taking an average or maximum over protected groups, both studied in this paper, are examples of post-processing methods. 

Much of the existing literature on bias takes data on outcomes as given, sometimes even referring to them as the ``ground truth,'' and formulates notions of bias that compare how well model predictions align with the training data for one group versus another. \citeA{criminal2021} discuss different statistical notions of fairness, and show that they can be impossible to jointly optimize.

By contrast, this paper considers the case in which the outcome data used to train the model include decisions that are biased against one group---e.g. loan applications that were denied despite the creditworthiness of the applicant. This approach explicitly distinguishes disparities in model predictions that owe to such biases in the training data from those arising from disparities in predictive variables.\footnote{\citeA{jiang2020id} study a similar problem, though they constrain the bias to solve a particular constrained optimization problem.} Even if one wants to minimize both, they may warrant very different policy responses. Equalizing denial rates for mortgage lending may be insufficient to close the racial wealth gap,\footnote{See \citeA{derenoncourt2021racial}.} for example. Furthermore, in the context of lending, giving members of a disadvantaged group loans they will be unable to repay may not improve the welfare of the group. Narrowing disparities in underlying indicators of creditworthiness may require policy interventions outside the scope of model de-biasing.

The US mortgage market is an important context, with \$1.8trn in new originations in 2018 (\citeR{cfpb2019}). Racial and ethnic disparities in mortgage credit access have been a source of concern for decades (see e.g.	\citeR{10.1257/jep.12.2.41}), and the stakes are indeed high: disparities in homeownership contribute to racial differences in wealth (\citeR{akbar2019racial}) and intergenerational wealth mobility (\citeR{toney2021intergenerational}). This paper contributes to the literature on discrimination in lending (see e.g. \citeR{bartlett2021consumer} and \citeR{zhang2021lenders}) as well as the literature in statistics and machine learning on model fairness and de-biasing (surveyed by e.g. \citeR{mehrabi2019survey}), showing that while machine learning models can replicate biased decisions in this context, the bias can (with some important caveats) be mitigated. The focus of this paper is methods of mitigating bias in underwriting models, leaving other fair lending concerns (such as bias  in pricing) to future work.

Section \ref{sec:data} describes the mortgage application context and the HMDA data used to study it. Section \ref{sec:disparities} describes several ways in which disparities can arise in model predictions between groups, even when sufficient data are available. Section \ref{sec:methods} describes some methods of de-biasing models to address these disparities. Section \ref{sec:results} compares the performance of each de-biasing method. 

\section{Context and Data}\label{sec:data}

The real-world context which we consider is modeling mortgage underwriting in the US. Mortgage applications include a variety of quantifiable data used to judge the creditworthiness of the application, including the loan-to-value ratio of the prospective loan, the borrower's debt-to-income ratio, and the borrower's credit score. Crucially for the purposes of this paper, the Home Mortgage Disclosure Act (HMDA) also requires that lenders ask for information about borrowers' race and ethnicity,\footnote{Specifically, ``Regulation C, 12 CFR $\S$ 1003.4(a)(10)(i), requires that a financial institution collect the ethnicity, race, and sex of a natural person applicant or borrower, and collect whether this information was collected on the basis of visual observation or surname.'' See https://www.consumerfinance.gov/policy-compliance/guidance/hmda-implementation/home-mortgage-disclosure-act-faqs/} though by law\footnote{Fair Housing Act (42 U.S.C. $\S\S$ 3601-19) and Equal Credit Opportunity Act (12 CFR $\S$ 1002).\label{fn:FHAECOA}} the decision to approve or deny a loan must not hinge on this prohibited basis. 

This paper uses HMDA mortgage application data from 2018. The data include the standard underwriting variables mentioned earlier as well as other information about the property such as its location, value, type, and construction method (the predictive variables used in the models are described in Section \ref{sec:model}). The sample is restricted to conventional\footnote{Guidelines for HMDA reporting define ``conventional'' as not insured or guaranteed by the FHA, VA, RHS, or FSA.} fully-amortizing 30-year home-purchase loans, excluding those for business or commercial purpose. This results in a sample of 2,067,993 applications, with an average denial rate of 6.5\%. In terms of prohibited basis groups, this paper considers two groups: applicants whose ethnicity is reported as Hispanic or Latino (7.2\% of the sample), and those of other ethnicity. 

\section{Sources of disparities in model predictions}\label{sec:disparities}

This section discusses several ways in which group disparities in predicted denial rates can emerge in this context, focusing on disparities unrelated to data availability or the ability of a model to approximate its training data, which are addressed elsewhere in the literature.\footnote{For example, \citeA{chen2018my} study the ability of data collection to ameliorate disparities in model predictions.} Throughout, assume that modelers have access to data and models capable of producing the expected value of an outcome given a set of predictive variables. The challenges addressed in this paper thus apply even when sample size is not a limitation.

Formally, let $y\in\{0,1\}^N$ be the outcomes of $N$ loan applications, where $y_i=1$ indicates that applicant $i$ was approved and $y_i=0$ denotes denial. Let $X_i$ be the characteristics of application $i$ available to use in a model (predictive variables, or features). Let $g_i$ denote $i$'s group, which may be a protected basis group such as a racial or ethnic group. Let $y^\text{fair}$ be a (possibly counterfactual) set of fair outcomes, which we can think of as the outcomes that would occur in the absence of lender bias.\footnote{Characterizing fair outcomes is beyond the aims of this paper; instead, we study the extent to which statistical de-biasing methods can eliminate bias relative to a known baseline.} Let $\hat{y}$ denote the modeler's (possibly de-biased) estimates of $y$ and $\epsilon$ denote residual variables that enter lenders' decisions but are not available for modeling, such that $y_i = \hat{y}_i + \epsilon_i$. Typically the modeler will specify a cutoff value $\underline{y}$ such that applicants with estimated $\hat{y}_i$ of at least $\underline{y}$ are predicted to be approved and the rest are predicted to be denied. 

\subsection{Explicit use of prohibited factors}\label{sec:disparatetreatment}
If some loan officers denied applications from a prohibited basis group (such as a racial or ethnic group) more often than justified by indicators of creditworthiness, then a model trained on their decisions may replicate that bias. Even if the variables used in the model do not include race or ethnicity, the model may use other variables (such as location or property type) to better match model predictions to the biased decisions. If some loan officers explicitly used prohibited basis group $g$ in their decisions, then $y_i<y^\text{fair}_i$ for some members of group $g$. As the model developers may be unaware which if any decisions were biased, studying methods that may prevent the model from replicating any such bias is the focus of this paper.\footnote{\citeA{discrimdc} provide direct evidence that some loan officers report perceived differences in creditworthiness between protected groups. \citeA{inacc2019} study inaccurate statistical discrimination.}

\subsection{Group differences in predictors}\label{sec:diffpredictors}
There may exist aggregate group differences in factors used by lenders to determine creditworthiness, such as credit score, employment history, and cash reserves. So even if lender decisions are not biased ($y=y^\text{fair}$), the indicators of creditworthiness on which they base their decisions may reflect upstream bias such as labor market discrimination or home appraisal bias, resulting in higher denial rates for one group than another.\footnote{\citeA{lewisfaupel2022hmda} describe such disparities in the HMDA data. \citeA{derenoncourt2021racial} trace the historical evolution of the racial wealth gap in the US. \citeA{bertrand2004emily} produce experimental evidence of discrimination in job search, and \citeA{tenev2018social} studies labor market inequality induced by differences in social networks.} In other words, the distribution of $X$ differs by group, and so $\text{E}\left(y\vert g\right)$ does as well. \footnote{When a facially neutral model disproportionately excludes certain persons on a prohibited basis, it may be described as having a disparate impact. While it is beyond the scope of this paper to comment on the legality of models used in lending or any other context, \citeA{FLhandbook} discusses other factors that may be relevant as well, ``including whether there is a robust causal link between the neutral policy or practice and the adverse effect(s) on members of a protected class and whether the policy or practice is necessary to achieve a legitimate business objective.'' See also 12 CFR $\S$ 1002.6.}

\subsection{Proxies for prohibited factors}\label{sec:proxy}
Another way in which disparities in model predictions might arise is if the loan officer decisions used to train the model were based on a proxy for a prohibited basis group, rather than group membership itself. For example, the loan officers may have denied applicants from a certain neighborhood because of the race or ethnicity of the people living there. A related concept is redlining, where a lender avoids certain neighborhoods because of the race or ethnicity of the people living there (\citeR{fedreg1994}). \citeA{aaronson2021holc} study some long-lasting effects of historical redlining, and contemporary examples exist as well (e.g. United States v. Cadence Bank, N.A. (N.D. Ga.), 2021). In this case individual outcomes may not depend on group conditional on predictive variables ($\text{E}\left(y\vert X,g\right)=\text{E}\left(y\vert X,g^\prime\right)$), but decisions are still biased in the sense that one group is more likely to receive unfairly adverse outcomes ($y_i<y^\text{fair}_i$) than the other. Discrimination by proxy is distinguished from explicit use of prohibited factors as the implications for de-biasing can be quite different, as Section \ref{sec:redlining} will show. Discrimination by proxy can also include the use of a predictive variable that is correlated with a protected group but of little use in predicting the outcome of interest.  

\subsection{Model selection bias}\label{sec:modelselect}
Given the differences in endowments and obstacles faced by different groups (see Section \ref{sec:diffpredictors}), it may be the case that the relationship between historical outcomes and predictive variables differs by group for reasons other than disparate treatment by lenders. For example, the relationship between loan-to-value ratio and approval may be weaker for a prohibited basis group if highly levered (high loan-to-value) applicants from that group tend to compensate by having more stable employment history (not recorded in the HMDA data, but typically observed by lenders).\footnote{A related concept is noise: \citeA{blattner2021noise} show that credit scores are noisier indicators of loan performance for minority groups than for others} Training a model without prohibited basis group as a predictive variable will often result in more accurate predictions for the majority group (see e.g. \citeR{chen2018my}). This may mean that the model predicts denial for some minority applicants even though their characteristics are not as predictive of denial for the minority group. Importantly, this can occur even given infinite data for all groups.

We define model selection bias as a model having less accurate predictions for one group than another because of the model selection process (e.g. optimizing overall accuracy). Our mortgage application data do exhibit characteristics indicative of this sort of model selection bias. For example, while Hispanic or Latino applicants have a higher denial rate overall and the chance of denial is generally increasing in debt-to-income (DTI) ratio for all groups, the denial rate for Hispanic or Latino applicants with high DTI ratios (above 80\%) is actually lower (75\%) than it is for other applicants (80\%). This suggests that the group with the highest denial rate at a given credit score, DTI, etc. may not have the highest denial rate for other values of those variables.

\section{De-biasing methods}\label{sec:methods}
This section describes several methods of de-biasing a model of approval decisions, including one (maximum over prohibited variables, Section \ref{sec:lop}) that is new to the literature. Section \ref{sec:results} will report empirical results for each method's success in removing counterfactual bias from mortgage application data. The methods considered here are only a subset of a growing literature, partially summarized in Section \ref{sec:intro}.

\subsection{Exclusion of prohibited variables}\label{sec:excluding}
A simple method of attempting a de-biased model is to simply exclude prohibited basis group from the predictors used to train the model: $\hat{y} = \text{E}\left(y\vert X\right)$. This method does provide predictions which do not depend explicitly on prohibited basis group (so it avoids explicit use of prohibited factors). However, \citeA{pope2011implementing} show that the exclusion of prohibited basis group from the list of predictors may result in the model overweighting other variables that are correlated with group, resulting in predictions that are more favorable to one group than another. Section \ref{sec:results} will demonstrate this empirically for the case of modeling mortgage approvals.

\subsection{Jointly optimizing accuracy and disparity between groups}\label{sec:regularization}
One popular method of de-biasing machine learning models is training the model to optimize not just accuracy alone but some weighted combination of accuracy and some measure of disparity between groups, sometimes called ``regularization.'' The specific method tested in this paper is FairXGBoost, proposed by \citeA{ravichandran2020fairxgboost}, which jointly optimizes the cross-entropy loss between the model predictions and the training data as well as the negative cross-entropy between the model predictions and the prohibited basis group.\footnote{Note that this is just one of many possible measures of disparity, and may not be an appropriate measure of fair lending risk.} This method thus balances the accuracy of its predictions against association between the prohibited basis group and the adverse outcome. However, such methods may fail to harness the full predictive power of variables associated with prohibited basis group, even if they are not directly related to the bias in the training data. Among the methods tested in this paper, only regularization requires the modeler to choose a parameter which determines the balance between disparity and accuracy.

\subsection{Averaging over prohibited variables}\label{sec:popesydnor}

Building on work by \citeA{ross2002color}, \citeA{pope2011implementing} propose a method for de-biasing regression models that can be adapted to the machine learning context as follows. If $X$ are acceptable predictive variables and $g$ are prohibited predictive variables, then define the averaging over prohibited variables method as:
\begin{equation}\label{eq:popesydnor}
	\text{E}_{\text{avg}}\left(y_i\vert X_i\right) \equiv \frac{1}{N}\sum_{j = 1}^N \text{E}\left(y_i\vert X_i,g_j\right).
\end{equation}

The main advantage of this method is that unlike regularization, it retains the within-group predictive power of the predictive variables. However, it can suffer from the model selection bias described in Section \ref{sec:modelselect}. If a prohibited basis group is a very small minority group, the de-biased predictions will simply approximate the expected outcomes for the majority group,\footnote{That is, $\text{E}\left(y\vert X\right)\sim\text{E}\left(y\vert X,g^{\text{majority}}\right)$.} and may fail to reflect the relationship between predictive variables and outcomes for the minority group.\footnote{Specifically, if $N_{g}$ is the size of group $g$ and $N_{g^\prime}$ is the size of group $g^\prime$, then $\lim_{N_{g}\rightarrow \infty}\frac{1}{N_g+N_{g^\prime}} \sum_{i = 1}^{N_g+N_{g^\prime}} \text{E}\left(y\vert X,g_i\right)$  $=\frac{1}{N_g} \sum_{i:g_i=g}^{N_g} \text{E}\left(y\vert X,g_i\right)$.} In particular, it may be the case that the de-biased model would deny a member of a prohibited basis group despite the modeler's best estimate ($\text{E}\left(y\vert X,g\right)$) being that the person is in fact creditworthy enough for approval. 

As machine learning models typically include interactions between predictive variables, unlike in the separable case considered by \citeA{pope2011implementing} there is no guarantee that Equation \ref{eq:popesydnor} will produce unbiased probabilities (that is, it may be that $\text{E}_{\text{avg}}\left(y\vert X\right)\neq\text{E}\left(y\vert X\right)$). In practice this can be addressed by scaling the predictions of the model.

\subsection{Maximum prediction over prohibited variable}\label{sec:lop}
To address the potential for model selection bias in method \ref{sec:popesydnor} as well as bias owing to explicit use of prohibited factors, consider the following novel form of de-biasing. A model (the ``full model'') is estimated using all available predictive variables, including prohibited basis group. Then, the debiased model predicts rejection for an applicant only if the full model predicts rejection for someone with those application characteristics regardless of which group they belong to.

Formally, max-over-groups prediction is defined as follows. If $\text{E}\left(y_i\vert X_i,g_i\right)$ is the expectation of outcome $y_i$ (where higher $y_i$ is favorable) for individual $i$ given observed characteristics $X_i$ and membership to group $g_i$, then the max-over-groups prediction is
\begin{equation}\label{eq:defn}
	\text{E}_{\text{max}} \left(y_i\vert X_i,g_i\right) \equiv \gamma \max_{g} \text{E}\left(y_i\vert X_i,g\right). 
\end{equation}

Here $\gamma$ is a scaling factor that can be set to 1 for un-scaled predictions or scaled to ensure that the ensuing predictions are not biased upwards for all prohibited basis groups.

This de-biasing method has the following intuitive properties. First, max-over-groups predictions are invariant to group (no explicit use of prohibited factors, as required by law\textsuperscript{\ref{fn:FHAECOA}} in lending).\footnote{That is, $\text{E}_{\text{max}} \left(y_i\vert X_i,g_i\right) = \text{E}_{\text{max}} \left(y_i\vert X_i,g_i^\prime\right)$ for $g_i\neq g_i^\prime$.} Second, anyone whose predicted approval in the full model hinges on prohibited basis group membership (all else constant) is approved in the de-biased model.\footnote{In other words, if the full model predicts approval for one applicant and rejection for another and prohibited basis is the only difference between the two applicants (i.e. they have the same credit score, debt-to-income ratio, etc.), both are approved in the de-biased model.} This can address both bias owing to explicit use of prohibited factors as well as model selection bias. The method is also easy to implement for data that include group membership. While this paper focuses on just two groups in the empirical section below, the method is easily implemented with multiple groups.

The max-over-groups predictions will have lower accuracy than a model which uses group as a predictor (though that is illegal in lending\textsuperscript{\ref{fn:FHAECOA}}), and will be biased upwards (favorable to applicants) for all individuals. They can be renormalized down to remove the overall bias, but this may actually result in wider disparities in denial rates compared to the group-blind model. Max-over-groups prediction is asymmetric in that it favors one outcome (approval) over the other. This may be to applicants' advantage in some cases, but may sometimes be a disadvantage---for example, if someone is approved for a loan they have trouble repaying.

Of the methods discussed above, only the first (which does not include prohibited variables) does not require data on prohibited group membership.\footnote{Some regularization methods use proxies for prohibited basis and thus do not require prohibited basis data for all instances---see e.g. \citeA{proxyfairness}.} Using prohibited basis group information for de-biasing purposes may present additional risks, legal and otherwise, depending on the context. This paper does not endorse any of the de-biasing methods studied, nor does it comment on the legality of these or any other methods which may use protected basis group data for de-biasing.

\section{Empirical Methods and Results}\label{sec:results}
\subsection{Random bias}\label{sec:random}
To evaluate the performance of each de-biasing method, counterfactual bias is added to mortgage decision data, randomly switching approvals to denials for Hispanic or Latino applicants such that the counterfactual denial rate for this group (19.1\%) is twice that observed in the data (9.5\%).  Decisions for applicants in other groups were left as in the original data. It is beyond the scope of this paper to evaluate the extent to which the mortgage underwriting decisions in the HMDA sample reflect bias.\footnote{For efforts in that direction, see e.g. \citeA{bhutta2022bias}} However, testing each de-biasing method's ability to recover the original decisions from the counterfactually biased data should be a useful test of performance.

\subsubsection{Excluding prohibited variables}\label{sec:model}
Next, this section demonstrates that a machine learning model trained on biased historical decision data can replicate that bias, even if the basis for the bias (e.g. ethnicity) is not used in the model, similar to \citeA{zhang2022bias}. A popular machine learning model, XGBoost, is trained to predict the counterfactually biased decisions (see Section \ref{sec:MLmodel} for details). 

The predictive variables (features) used are credit score (the lowest, if more than one applicant), combined loan-to-value ratio, debt-to-income ratio, income, property value, construction method, property type, total number of units, state, and county. More detail on these variables and the HMDA data in general is available from the \citeA{ffiec}. 

\subsubsection{Other de-biasing methods}
The other de-biasing methods are then tested as follows. FairXGBoost is trained on the counterfactually biased data with a parameter value (the weight given to reducing association between the outcome and protected group) of 0.2. While this choice is somewhat arbitrary, in practice the original decisions will not be available to tune this parameter. 

The average-over-groups method is executed by first training an XGBoost model including the prohibited basis group indicator (Hispanic/Latino). A prediction for each applicant is then generated as follows. 500 instances of prohibited basis group are drawn at random from the population, and a prediction is generated for the applicant assuming they belong to that group; the ultimate de-biased prediction is the average over these 500 predictions. In the case of two groups this is essentially equivalent to a simple weighted average (with population weights) of the prediction for a Hispanic applicant and the prediction for a non-Hispanic applicant with otherwise identical characteristics, but see Section \ref{sec:redlining} to see how this method works in a more complicated case.

Finally, to test the de-biasing power of max-over-groups prediction in this context, an XGBoost model is trained using the same predictive variables, again including Hispanic or Latino ethnicity. Then, two predictions are generated for each individual: the first assuming the individual is Hispanic or Latino, and the second assuming they are not. The maximum over groups prediction (hereafter, ``max-over-groups'', or ``max'') for that individual is then whichever of these two predicted approval probabilities is higher.

In terms of accuracy, how do the de-biased models compare? Table \ref{tab:AUC} shows the AUCs (area under the receiver operating characteristic curve) for each de-biasing method, all trained on the counterfactually biased data. The ``Biased'' row for each method shows how well the method predicts biased data, giving the AUCs for that method's predictions compared to the counterfactually biased decision data (both on the held-out test sample). The ``Actual'' row (in blue) gives the AUCs comparing the predictions to the original data, showing how well the method is able to recover the original decisions. 

Of the de-biasing methods, taking the maximum or average prediction over PBGs perform very similarly and provide the highest AUCs compared to the actual data. The Exclusion method performs nearly as well. Furthermore, these methods perform as well (in terms of AUC with respect to the actual data) as an XGBoost model trained on all actual data, including PBG indicators (results not shown in this table). The FairXGBoost regularization method has the least predictive power, as it sacrifices predictive power to reduce the association between the predicted outcomes and the protected group.  However, note that these and subsequent comparisons lack confidence intervals to give a sense of how confident we should be in the rank ordering.

\begin{table}[H]

	\bigskip{}
	\begin{onehalfspace}
		\begin{tabular}{ccccc}
			\multicolumn{1}{c}{} & & (1) & (2) & (3) \tabularnewline
			\multicolumn{1}{c}{} De-biasing method & Compared to & Hispanic/Latino & Not Hispanic/Latino & All \tabularnewline
			\hline 
			\hline
			\multicolumn{1}{c}{} None & Biased & 0.625 & 0.785 & 0.798 \tabularnewline
			\multicolumn{1}{c}{}  & \textcolor{blue}{Actual} & \textcolor{blue}{0.765} & \textcolor{blue}{0.785} & \textcolor{blue}{0.780}
			\tabularnewline
			\multicolumn{1}{c}{} Exclude PBG & Biased & 0.626 & 0.784 & 0.776 \tabularnewline
			\multicolumn{1}{c}{}  & \textcolor{blue}{Actual} & \textcolor{blue}{0.765} & \textcolor{blue}{0.784} & \textcolor{blue}{0.784}
			\tabularnewline
			\multicolumn{1}{c}{} FairXGB & Biased & 0.585 & 0.706 & 0.688 \tabularnewline
			\multicolumn{1}{c}{}  & \textcolor{blue}{Actual} & \textcolor{blue}{0.676} & \textcolor{blue}{0.706} & \textcolor{blue}{0.704}
			\tabularnewline
			\multicolumn{1}{c}{} Avg. over PBG & Biased & 0.626 & 0.784 & 0.770 \tabularnewline
			\multicolumn{1}{c}{}  & \textcolor{blue}{Actual} & \textcolor{blue}{0.770} & \textcolor{blue}{0.784} & \textcolor{blue}{0.785}
			\tabularnewline
			\multicolumn{1}{c}{} Max. over PBG & Biased & 0.627 & 0.785 & 0.770 \tabularnewline
			\multicolumn{1}{c}{}  & \textcolor{blue}{Actual} & \textcolor{blue}{0.769}  & \textcolor{blue}{0.785}  & \textcolor{blue}{0.785}
			\tabularnewline
			\hline
		\end{tabular}
	\end{onehalfspace}
\scriptsize{\emph{Note: Models trained on approval data including random counterfactual denials of Hispanic or Latino applicants. AUCs compare de-biased predictions on held-out test sample to outcomes with counterfactual bias (rows marked ``Biased'') and without (rows marked ``Actual'').}}
\caption{AUC \label{tab:AUC}}
\end{table}

The XGBoost model generates a score corresponding to the likelihood that each application would have been denied (including the counterfactually biased denials). The AUC results above measure accuracy of predictions across all possible thresholds for approval. To see how each method performs for a specific, realistic cutoff threshold, assume that the lender sets a cutoff score for denial which results in a predicted denial rate that matches the overall denial rate in the biased training data, 7.2\%. 

Table \ref{tab:denial} shows the denial rates for the predictions using each de-biasing method and the threshold described above, as well as the denial rates in the biased data used to train the models (penultimate row) and the actual data from which the biased data were derived (last row). By design, the overall denial rates for each method (right column) very nearly match the overall denial rate in the training data, with some slight error coming from the fact that the denial rates for the predictions were computed on the held-out test subsample of the data rather than the training subsample. 

\begin{table}[H]

	\bigskip{}
	\begin{onehalfspace}
		\begin{tabular}{cc|ccc}
			\multicolumn{1}{c}{} & & Hispanic/Latino & Not Hispanic/Latino & All \tabularnewline
			\hline 
			\hline
			 & Exclude PBG & 17.3 & 6.4 & 7.1\tabularnewline
			 Predictions & FairXGB & 10.9 & 6.9 & 7.2\tabularnewline
			 & Avg. over PBG & 13.5 & 6.7 & 7.2\tabularnewline
			 & Max. over PBG & 13.6 & 6.7 & 7.2\tabularnewline
			\hline
			 & Biased training data & 19.3 & 6.2 & 7.2 \tabularnewline
			 & Actual data & 9.5 & 6.2 & 6.5 \tabularnewline
			\hline 
		\end{tabular}
	\end{onehalfspace}
\caption{Denial rates\label{tab:denial}}
\end{table}

Simply excluding the PBG (ethnicity indicators) from the training data is insufficient to remove the counterfactual bias: the predicted denial rate for Hispanic or Latino applicants using this method is nearly as high as in the biased training data. FairXGBoost results in the smallest raw disparity in predicted denial rates between the two groups. However, it is unclear from looking at raw denial rates whether this comes from correctly identifying applicants who were counterfactually denied.

\begin{figure}[h]
	\centering
	\begin{minipage}{\linewidth}
		\includegraphics[trim=0 0 0 0, clip, scale=0.63]{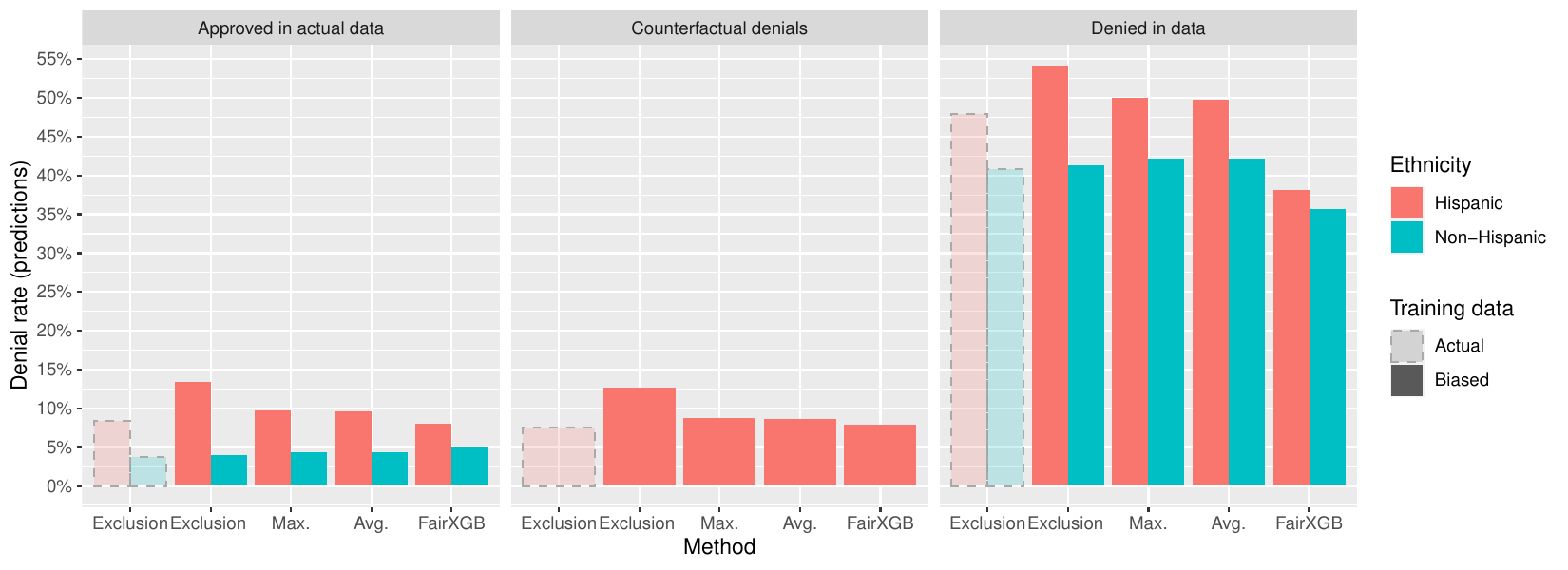}
		\begin{flushleft}
			\scriptsize{\emph{Note: Biased training data includes random counterfactual denials of Hispanic/Latino applicants. Apart from FairXGB, the different de-biasing methods are labeled according to how they treat the PBG, ethnicity: ``Exclusion'' excludes the PBG from the set of predictive features; ``Maximum'' takes the maximum prediction over different PBGs; ``Average'' takes the average prediction over PBGs.}}
		\end{flushleft}
	\end{minipage}
\caption{Predicted denial rates by actual disposition, model, and ethnicity\label{fig:random}}
\end{figure}

Figure \ref{fig:random} shows, for each de-biasing method, the denial rate by PBG and disposition in the original data. The left panel shows denial rates for those applications that were approved in the original data. The height of each bar shows the denial rate for each de-biasing method (on the horizontal axis) and ethnicity (denoted by the color of the bar). The middle panel shows denial rates for the counterfactual denials---applications that were approved in the original data, but labeled ``denied'' in the training data to simulate bias. The right panel shows applications denied in both the original data and the biased training data. 

In this visualization, a well-performing model will have low denial rates for the first two panels (which were approved in the actual data) and high denial rates for the third. The faded bars with dashed borders marked Exclusion (the leftmost within each panel) give denial rates for a model trained on the actual data (though still excluding PBG as a predictor), which gives a benchmark against which to measure the de-biasing performance of the models trained on the biased data.

Here we see that excluding the PBG from the set of predictors results in the highest denial rates for Hispanic/Latino applicants regardless of their disposition in the original data, emphasizing that this de-biasing method is insufficient to prevent biased training data from resulting in biased predictions. Note that the model predictions for the Exclusion method are perfectly explained by factors other than prohibited basis group. However, it would be erroneous to conclude from this (in the context of a fair lending exercise, for example) that either the model or the previous predictions are unbiased. 

FairXGBoost regularization results in the fewest denials of Hispanic/Latino applicants who were approved in the actual data. However, this comes at the cost of identifying the fewest actual denials (right panel of Figure \ref{fig:random}), and denying more approved non-Hispanic/Latino applicants than any other model (left-most panel). In this case, lower accuracy means a denial rate closer to average regardless of the disposition in the training data. 

Taking the maximum approval prediction over prohibited basis groups results in the next-lowest denial rate for the counterfactual denials, with similar accuracy in identifying actual denials to the average-over-PBG method. This demonstrates that these methods are able to recover some original decisions from biased data, while still preserving predictive power similar to that of a model trained on the original data (represented by the faded bars at the left of each panel).

One benefit of taking the maximum over groups rather than the average is the ability to address model selection bias of the form discussed in Section \ref{sec:modelselect}, by approving applicants from minority groups in regions of the feature space in which they tend to have better outcomes than the majority. In this exercise, however, the counterfactual bias mostly precludes any opportunity for correcting this sort of model selection bias. For 97\% of the sample Hispanic or Latino applicants, the max.-over-groups prediction assigned is the non-Hispanic/Latino prediction. So in this case, the method gives almost everyone a prediction as if they belonged to the majority group. This would not be the case with less bias, however---when a model is trained on the actual data (rather than the biased data), the max.-over-groups method assigns the Hispanic/Latino prediction to 18\% of Hispanic or Latino applicants, demonstrating that model selection bias is indeed a concern.

\subsection{Bias based on location}\label{sec:redlining}

Section \ref{sec:random} showed that de-biasing methods can remove bias that takes the form of random adverse decisions against a particular group. Historically, however, bias against a particular group can alternatively be targeted through a proxy---variables correlated with group membership, such as location (see Section \ref{sec:proxy}). This section shows that the success of de-biasing methods depends crucially on how bias was targeted: in the case of bias based on property location, some methods can (without additional modification) actually result in worse disparities than the baseline model.

A new set of biased training data is generated as follows. Applications in the census tracts with the highest proportions of Hispanic or Latino residents are counterfactually switched to denial until the total number of counterfactual denials matches that in Section \ref{sec:random}. This simulates location bias: denial of applications based on predominant neighborhood ethnicity. Since these counterfactual denials target applicants based on geography, they affect non-Hispanic or Latino applicants as well (though proportionally much less).

Next, as before, an XGBoost model is trained on the counterfactually biased data and each de-biasing method is tested. In this case, for the average-over-groups method the average is taken over both ethnicity indicators as well as the geography variables (county indicators). In other words, 
\begin{equation}\label{eq:popesydnorgeo}
	\text{E}_{\text{avg}}\left(y_i\vert X_i\right) \equiv \frac{1}{N}\sum_{j = 1}^N \text{E}\left(y_i\vert X_i,g_j,\text{county}_j\right)
\end{equation}
where in this case $X$ is all the predictive variables excluding ethnicity and county. If the modeler is aware that the data may be biased based both on ethnicity and geography, they can include these variables in the list of prohibited variables that are averaged over, at the cost of sacrificing their predictive power.  

The max.-over-groups method will still take the most favorable prediction over just the two ethnicity groups as before. Since there are so many combinations of county/ethnicity in the data, taking the most favorable one would likely result in approval for all applicants---an unhelpful result. This highlights an important limitation of the max.-over-groups method: it may not be useful when there are a large number of prohibited variables. The exclusion method will exclude ethnicity but not county, to demonstrate the insufficiency of narrowly focusing on only the protected group but ignoring likely proxies.

\begin{figure}[h]
	\centering
	\begin{minipage}{\linewidth}
		\includegraphics[scale=0.63, trim=0 0 0 0, clip]{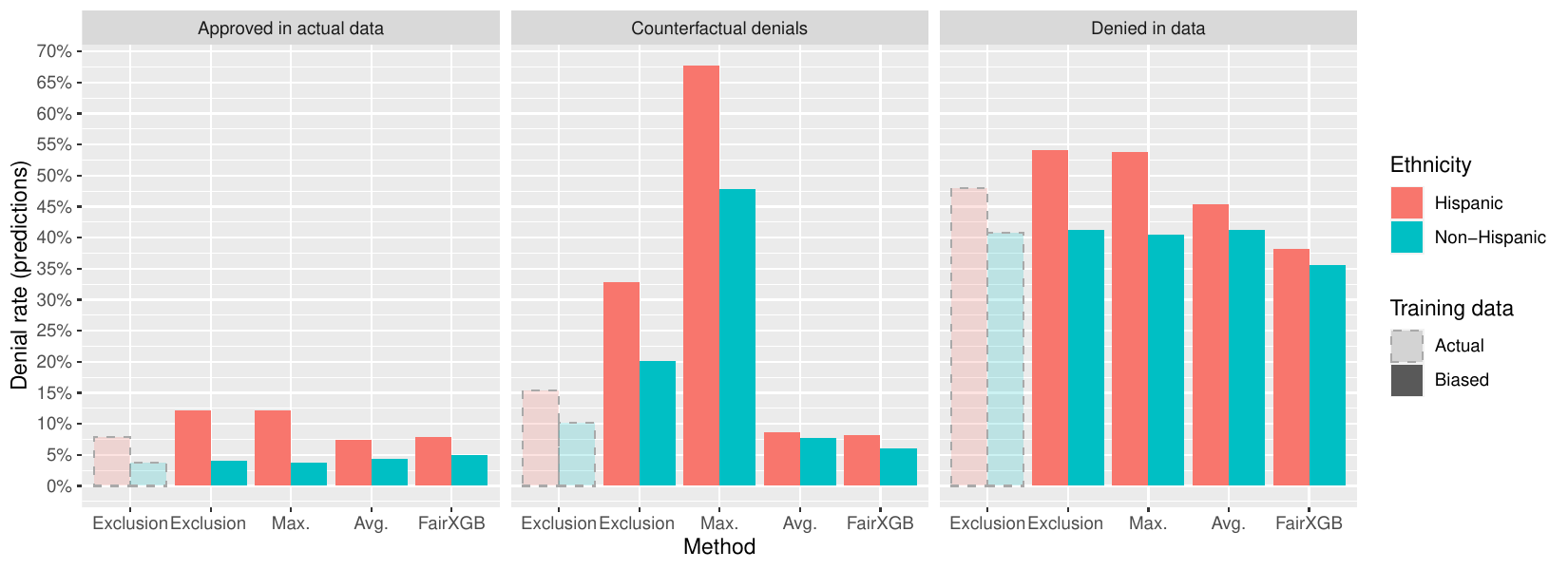}
		\begin{flushright}
			\scriptsize{\emph{Note: Biased training data includes counterfactual denials of applicants in census tracts with high Hispanic populations.}}
		\end{flushright}
	\end{minipage}
\caption{Predicted denial rates by actual disposition, model, and ethnicity\label{fig:redlining}}
\end{figure}

Figure \ref{fig:redlining} shows, for the models trained on the simulated location bias data, the denial rates for each de-biasing method and disposition in the original data. As before, the thresholds for denial are set such that the average denial rates match that of the training sample. 

For the simulation of bias based on location, excluding PBG alone is again insufficient to remove the bias, resulting in high denial rates for applicants who were counterfactually denied (middle panel). This is true despite the bias being enacted at a level (census tract) finer than the geographic variables included as predictors in the model (county). The FairXGBoost method predicts few denials for those approved in the actual data, but is again worse than the other methods at identifying those who were actually denied (right panel). 

The max.-over-groups method is noticeably worse at removing bias based on location than it was at removing direct bias in the training data. This method fails to remove $\frac{2}{3}$ of the counterfactual denials of Hispanic or Latino applicants (middle panel), and replicates nearly half of the counterfactual denials of other applicants. Intuitively, the max.-over-groups removes bias by giving applicants the most favorable outcome observed in a particular region of the feature space. Since location bias affects members of all groups in certain areas, the lack of approvals in areas with high concentrations of Hispanic or Latino residents prevents this method from effectively removing the bias.

By contrast, the average-over-groups method performs much better, resulting in low denial rates for approved and counterfactually denied applicants while maintaining reasonably high denial rates for those actually denied in the original data. This demonstrates that if modelers are able to identify variables that might have been used to enact bias, it can be effectively mitigated by averaging over those variables.

\section{Conclusion}

Practitioners often want to automate decisions, but if a model is trained to replicate decisions that are biased against a certain group then the model may replicate that bias. This paper uses counterfactually biased mortgage decision data to empirically test several methods of mitigating such bias, including adapting one to a machine learning context (averaging over prohibited variables \`a la \citeR{pope2011implementing}) and introducing another (max.-over-groups) that can mitigate bias by treating applicants as if they belong to the group that would result in the most favorable outcome for them. 

Excluding prohibited basis group from the list of predictors is shown to be an insufficient form of bias mitigation, as a sufficiently flexible model can find proxies with which to replicate the bias. Regularization that jointly optimizes disparity and accuracy can reduce inter-group disparities in model predictions, but at a cost of accuracy. This may be counterproductive in some contexts such as lending, where being approved for a loan one cannot repay may be worse than denial. Taking the average or most favorable prediction over prohibited groups can mitigate bias against a protected group, and the latter can mitigate model selection bias as well, rewarding minority applicants who are more creditworthy than others with similar characteristics. But when bias against a group is enacted via a proxy, averaging over both protected group and the problematic predictive variable may better mitigate the bias.\footnote{As noted before, this paper does not comment on the legality of collecting or using protected group information in training or de-biasing models.} These results highlight the importance of context: understanding the forms that bias can take can be essential to its mitigation.

\acks The author is grateful to participants in the OCC Economics Seminar Series and colleagues in the Compliance Risk Analysis Division for valuable comments and advice.

\bibliography{bias}
\bibliographystyle{theapa}

\section{Appendix}
\subsection{Machine learning model details}\label{sec:MLmodel}
XGBoost, a scalable end-to-end tree boosting system (\citeR{xgboost}) was implemented in R. The hyperparameters listed in Table \ref{tab:hyperparameters} were found via 5-fold cross-validation to produce the maximum test AUC of 0.798 (see Table \ref{tab:AUC}) on the counterfactually biased data (using no de-biasing), across a grid of possible combinations of parameter values. Table \ref{tab:hyperparameters} gives the grid of hyperparameters that were tried, as well as the final values chosen to produce the maximum test AUC. 
\begin{table}[H]
	\caption{Hyperparameters chosen by cross-validation\label{tab:hyperparameters}}
	\begin{onehalfspace}
				\centering{}%
		\begin{tabular}{c|ccc|c}
			& & Grid && \tabularnewline
			& Min & Step & Max & Value chosen \tabularnewline
			\hline
			Number of rounds & 1 & 1 & 500 & 363 \tabularnewline
			$\eta$ & 0 & 0.1 & 0.3 & 0.2 \tabularnewline
			Maximum depth & 2&2&10& 8 \tabularnewline
			Minimum child weight & 1 & $5^x$ & 625 & 25 \tabularnewline
			\hline 
		\end{tabular}
	\end{onehalfspace}
\end{table}

Other XGBoost settings were left at default values. These hyperparameters were used for all models. Each model was trained on a 80\% sample, and all reported results (Tables \ref{tab:AUC} and \ref{tab:denial}) refer to model predictions on a held-out 20\% sample.

\end{document}